\pdfoutput=1
\documentclass[11pt]{article}

\usepackage{emnlp2021}

\usepackage{times}
\usepackage{latexsym}
\usepackage{booktabs}
\usepackage{array}
\usepackage{subfigure}
\usepackage{pgfplots}
\usepackage{arydshln}
\usepackage{amsmath}
\usepackage{xcolor}
\usepackage[T1]{fontenc}

\usepackage[utf8]{inputenc}

\usepackage{microtype}

\definecolor{q1}{HTML}{2ca02c}
\definecolor{q2}{HTML}{9467bd}
\definecolor{q3}{HTML}{1f77b4}
\definecolor{q4}{HTML}{8c564b}
\definecolor{q5}{HTML}{d62728}
\definecolor{q6}{HTML}{ff7f0e}
\title{Mixture-of-Partitions:\\ Infusing Large Biomedical Knowledge Graphs into BERT}

\author{Zaiqiao Meng$^\spadesuit$, Fangyu Liu$^\spadesuit$, Thomas Hikaru Clark$^\spadesuit$\\ \textbf{Ehsan Shareghi}$^\clubsuit$$^\spadesuit$, \textbf{Nigel Collier}$^\spadesuit$ \\
$^\spadesuit$Language Technology Lab, University of Cambridge \\
$^\clubsuit$Department of Data Science and AI, Monash University\\

  \texttt{$^\spadesuit$\{zm324, fl399, thc44, nhc30\}@cam.ac.uk} \\
  \texttt{$^\clubsuit$ehsan.shareghi@monash.edu}
  }

\begin{document}
\maketitle
\begin{abstract}
Infusing factual knowledge into pretrained models 
is fundamental for many knowledge-intensive tasks. In this paper, we propose \textbf{Mixture-of-Partitions (MoP)}, an 
infusion approach that can handle a very large knowledge graph (KG) by partitioning it into smaller sub-graphs and infusing their specific knowledge into various BERT models using lightweight adapters. To leverage the overall factual knowledge for a target task, these sub-graph adapters are further fine-tuned along with the underlying BERT through a mixture layer.
We evaluate our MoP with three biomedical BERTs (SciBERT, BioBERT, PubmedBERT) on six downstream tasks (inc. NLI, QA, Classification), and the results show that our MoP consistently enhances the underlying BERTs in task performance, and achieves new SOTA performances on five evaluated datasets.
\footnote{Our code, models and other related resources can be found in \url{https://github.com/cambridgeltl/mop}.}



\end{abstract}

\section{Introduction}

\begin{figure*}[t]
	\centering
	\includegraphics[width = 155mm]{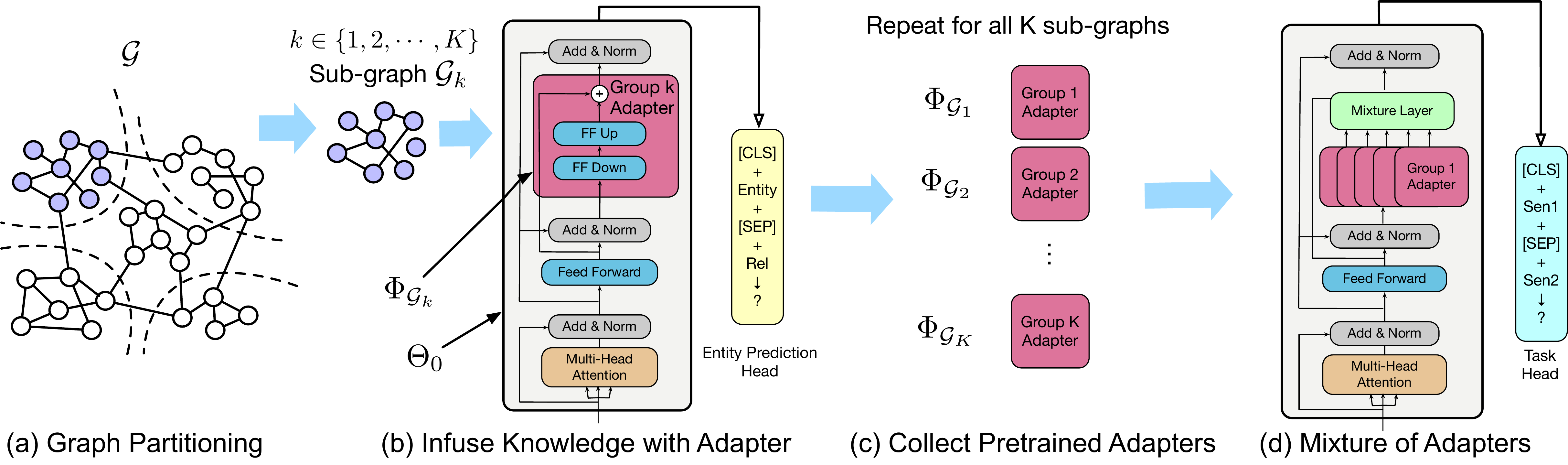}
	\caption{Overview of the proposed MoP.}
	\label{fig:model}
\end{figure*}
Leveraging
factual knowledge to augment pretrained language models 
is of paramount importance for knowledge-intensive tasks, such as question answering and fact checking~\cite{petroni2020kilt}.  Especially in the biomedical domain where public training corpora are limited and noisy, trusted biomedical KGs are crucial for deriving accurate inferences~\cite{li2020towards,liu2020self}. However, the infusion of knowledge from real-world biomedical KGs, where entity sets are very large (e.g. UMLS, \citealt{bodenreider2004unified}, contains $\sim$4M entities) demands highly scalable solutions.



Although many general knowledge-enhanced language models have been proposed, most of them rely on a computationally expensive joint training of an underlying masked language model (MLM) along with a knowledge-infusion objective function  to minimize the risk of catastrophic forgetting~\cite{xiong2019pretrained,zhang2019ernie,wang2019kepler,wang2020k,peters2019knowledge,yuan2021improving}.
Alternatively, entity masking (or entity prediction) has emerged as one of the most popular self-supervised training objectives for infusing entity-level knowledge into pretrained models~\cite{sun2019ernie,zhang2019ernie,yu2020jaket,he2020infusing}. However, due to the large number of entities in biomedical KGs, computing an exact softmax over all entities is very expensive for training and predicting~\cite{de2020autoregressive}. Although negative sampling techniques could alleviate the computational issue~\cite{sun2020colake}, tuning an appropriately hard set of negative instances can be challenging and predicting a very large number of labels may generalize poorly~\cite{hinton2015distilling}.

To address the aforementioned challenges, we propose a novel knowledge infusion approach, named \textbf{Mixture-of-Partitions (MoP)}, to infuse 
factual knowledge based on partitioned KGs into pretrained models (BioBERT, \citealt{lee2020biobert}; SciBERT, \citealt{beltagy2019scibert}; and PubMedBERT, \citealt{gu2020domain}). More concretely, we first partition a KG into several sub-graphs each containing a disjoint subset of its entities by using the \textsc{METIS} algorithm~\cite{karypis1998fast}, and then the Transformer \textsc{Adapter} module ~\cite{houlsby2019parameter,pfeiffer2020adapterhub} is applied to learn portable knowledge parameters from each sub-graph. 
In particular, using \textsc{Adapter} module to infuse knowledge does not require fine-tuning the parameters of the underlying BERTs, which is more flexible and efficient while avoiding the catastrophic forgetting issue. To utilise the independently learned knowledge from sub-graph adapters,
we introduce mixture layers to automatically route useful knowledge from these adapters to downstream tasks. Figure~\ref{fig:model} illustrates our approach. 

Our results and analyses indicate that our ``divide and conquer" partitioning strategy effectively preserves the rich information presented in two biomedical KGs from {UMLS} while enabling us to scale up training on these very large graphs. Additionally, we observe that while individual adapters specialize towards sub-graph specific knowledge, MoP can effectively utilise their individual expertise to enhance the performance of our tested biomedical BERTs on six downstream tasks, where five of them achieve new SOTA performances.

\section{Mixture-of-Partitions (MoP)}

We denote a KG as a collection of ordered triples $\mathcal{G}=\{(h,r,t) | h,t \in \mathcal{E}, r \in\mathcal{R}\}$, where $\mathcal{E}$ and $\mathcal{R}$ are the sets of entities and relations, respectively. All the entities and relations are associated with their textual surface forms, which can be a single word (e.g. \emph{fever}), a compound (e.g. \emph{sars-cov-2}), or a short phrase (e.g. \emph{has finding site}).

Given a pretrained model $\Theta_{0}$, 
our task is to learn 
$\Phi_{\mathcal{G}}$ based on an input knowledge graph  $\mathcal{G}$,
such that it encapsulates the knowledge from $\mathcal{G}$. The training objective, $ \mathcal{L}_{\mathcal{G}}$, can be implemented in many ways such as relation classification~\cite{wang2020k}, entity linking~\cite{peters2019knowledge}, next sentence prediction~\cite{goodwin2020enhancing}, or entity prediction~\cite{sun2019ernie}. In this paper, we focus on entity prediction, one of the most widely used objectives, and leave exploration of other objectives for future work. 

As mentioned earlier, exact softmax over all entities is extremely expensive~\cite{mikolov2013distributed,de2020autoregressive} for large-scale KGs, hence we resort to the principle of ``divide and conquer'', and propose a novel approach called {Mixture-of-Partition (MoP)}. Specifically, our MoP first partitions a large KG into smaller sub-graphs (i.e., $\mathcal{G}\rightarrow\{\mathcal{G}_{1},\mathcal{G}_{2},\dots, \mathcal{G}_{K}\}$, \S\ref{subsec:partition}), and learns sub-graph specific parameters on each sub-graph separately (i.e., $\{\Phi_{\mathcal{G}_1},\Phi_{\mathcal{G}_2},\dots,\Phi_{\mathcal{G}_K}\}$, \S\ref{sec:adapters}).  Then these sub-graph parameters are fine-tuned through {mixture layers} to route the sub-graph specific knowledge into a target task (\S\ref{sec:mixturelayers}). 

\begin{table*}[t]
	\centering
	\vspace{-1em}
	\setlength{\tabcolsep}{5.2pt}
	\small 
		\begin{tabular}{lccccccccc}
			\toprule
			\textbf{Model$\downarrow$, Dataset$\rightarrow$} & \textbf{HoC} & \textbf{PubMedQA} &  \textbf{BioASQ7b} & \textbf{BioASQ8b}  &  \textbf{MedQA} & \textbf{MedNLI} \\
			\midrule
			\textbf{SciBERT}  & 80.52$_{\pm0.60}$\;\; & 57.38$_{\pm4.22}$\;\; & 75.93$_{\pm4.20}$\;\; & 75.72$_{\pm1.79}$\;\; & 29.42$_{\pm0.94}$\;\; & 81.19$_{\pm0.54}$\;\; \\
			\;\; + \textbf{MoP (SFull)} & 81.48$^\dagger_{\pm0.35}$$\uparrow$ & 54.78$_{\pm2.96}$\;\; & 79.14$^\dagger_{\pm3.27}$$\uparrow$ & 74.74$_{\pm1.96}$\;\; & 33.03$^\dagger_{\pm0.72}$$\uparrow$ & 81.43$_{\pm0.34}$$\uparrow$\\
			\;\; + \textbf{MoP (S20Rel)} & 81.79$^\dagger_{\pm0.66}$$\uparrow$ &54.66$_{\pm3.10}$\;\; &78.50$^\dagger_{\pm4.06}$$\uparrow$ &76.25$_{\pm2.20}$$\uparrow$ &32.77$^\dagger_{\pm0.67}$$\uparrow$ &81.20$_{\pm0.37}$\;\;
			&\\\hdashline
			\textbf{BioBERT}  & 81.41$_{\pm0.59}$\;\; & 60.24$_{\pm2.32}$\;\; & 77.50$_{\pm2.92}$\;\; & 78.75$_{\pm4.16}$\;\; &  30.48$_{\pm0.55}$\;\; & 82.42$_{\pm0.59}$\;\; & \\
			\;\; + \textbf{MoP (SFull)}  & 81.47$_{\pm0.89}$$\uparrow$ & 61.82$^\dagger_{\pm1.04}$$\uparrow$ & 81.29$^\dagger_{\pm3.46}$$\uparrow$& 82.04$^\dagger_{\pm4.59}$$\uparrow$ & 33.55$^\dagger_{\pm0.56}$$\uparrow$ & 83.44$_{\pm0.24}$$\uparrow$\\
			\;\; + \textbf{MoP (S20Rel)} & 82.53$^\dagger_{\pm1.08}$$\uparrow$ &61.04$_{\pm4.81}$$\uparrow$ &80.79$^\dagger_{\pm4.40}$$\uparrow$ &80.00$_{\pm5.03}$$\uparrow$ & 34.15$^\dagger_{\pm0.79}$$\uparrow$ &82.93$_{\pm0.55}$$\uparrow$ &\\\hdashline
			\textbf{PubMedBERT}& 82.25$_{\pm0.46}$\;\; & 55.84$_{\pm1.78}$\;\; & 87.71$_{\pm4.25}$\;\; & 84.54$_{\pm2.36}$\;\; & 35.08$_{\pm0.22}$\;\; & 84.18$_{\pm0.19}$\;\;\\
			\;\; + \textbf{MoP (SFull)}  & 82.71$_{\pm0.56}$$\uparrow$ & 61.74$^\dagger_{\pm2.54}$$\uparrow$ & 88.64$_{\pm3.04}$$\uparrow$ & {86.12}$^\dagger_{\pm2.39}$$\uparrow$ & {36.33}$^\dagger_{\pm0.16}$$\uparrow$ & 84.25$_{\pm0.25}$$\uparrow$\\
			\;\; + \textbf{MoP (S20Rel)} & \textbf{83.26}$^\dagger_{\pm0.32}$$\uparrow$ & \textbf{62.84}$^\dagger_{\pm2.71}$$\uparrow$ & \textbf{90.64}$^\dagger_{\pm2.43}$$\uparrow$ & 85.39$_{\pm1.51}$$\uparrow$ & \textbf{38.02}$^\dagger_{\pm0.05}$$\uparrow$ & \textbf{84.70}$_{\pm0.19}$$\uparrow$\\
			\bottomrule
			\textbf{SOTA} &  \vtop{\hbox{\strut \;{82.32}}\hbox{\strut \scriptsize \cite{gu2020domain}}} & \vtop{\hbox{\strut \;{60.24}}\hbox{\strut \scriptsize \cite{gu2020domain}}} & \vtop{\hbox{\strut \;{87.56}}\hbox{\strut \scriptsize \cite{gu2020domain}}} & 
			\vtop{\hbox{\strut \;\;\;\;\textbf{90.32}}\hbox{\strut \scriptsize \cite{nentidis2020overview}}} & \vtop{\hbox{\strut \;{36.70}}\hbox{\strut \scriptsize \cite{jin2020disease}}} & \vtop{\hbox{\strut \;\;83.80}\hbox{\strut \scriptsize \cite{peng2019transfer}}} \\
			\bottomrule
		\end{tabular}
	\caption{Performance on various tasks. The best ones are in \textbf{bold}, while $\uparrow$ denotes that improvements are observed comparing with the base model. ``$^\dagger$'' denotes statistically significant better than the base model (T-test, $p<0.05$).}
	\label{tab:performance}
		\vspace{-1em}
\end{table*}
\subsection{Knowledge Graph Partitioning}
\label{subsec:partition}
Graph partitioning (i.e., partitioning the node set into mutually exclusive groups) is a critical step to our approach, since we need to properly and automatically cluster knowledge triples for supporting data parallelism and controlling computation.
In particular, it must satisfy the following goals: (1) maximize the number of resulting knowledge triples to retain as much factual knowledge as possible; (2) balance nodes over partitions to reduce the overall parameters across different entity prediction heads; (3) efficiency at scale for handling large KGs. In fact, an exact solution to (1) and (2) is referred to as the \em balanced graph partition \em problem, which is NP-complete. We use the \textsc{METIS}~\cite{karypis1998fast} algorithm as an approximation, simultaneously meeting all the above three requirements. \textsc{METIS} can handle billion-scale graphs by successively coarsening a large graph into smaller graphs, processing them quickly and then projecting the partitions back onto the larger graph, and has been used in many tasks~\cite{chiang2019cluster,defferrard2016convolutional,zheng2020distdgl}. 

\subsection{Knowledge Infusion with Adapters}
\label{sec:adapters}
Once the large knowledge graph is partitioned, we use \textsc{Adapter} modules to infuse the factual knowledge into a pretrained Transformer model by training an entity prediction objective for each sub-graph. \textsc{Adapter}s~\cite{houlsby2019parameter,pfeiffer2020adapterhub} are newly initialized modules inserted between the Transformer layers of a pretrained model. The training of \textsc{Adapter} does not require fine-tuning the existing parameters of the pretrained model. Instead, only the parameters within the \textsc{Adapter} modules are updated. In this paper, we use the \textsc{Adapter} module configured by \citet{pfeiffer2020adapterfusion}, which is shown in Figure~\ref{fig:model}~(b). In particular, given a sub-graph $\mathcal{G}_k$, we remove the tail entity name for each triple $(h,r,t)\in \mathcal{G}_{k}$, and transform the triple into a list of tokens: `[CLS] $h$ [SEP] $r$ [SEP]'. The sub-graph specific \textsc{Adapter} module is trained to predict the tail entity using the representation of the [CLS] token and the parameters $\Phi_{\mathcal{G}_k}$ are optimized by minimizing the cross-entropy loss. During the fine-tuning of downstream tasks, both the parameters of \textsc{Adapter} and pre-trained LM will be updated. 
\subsection{Mixture Layers}
\label{sec:mixturelayers}

Given a set of knowledge-encapsulated adapters, 
we use {AdapterFusion} mixture layers to combine knowledge from different adapters for downstream tasks. {AdapterFusion} is a recently proposed  model~\cite{pfeiffer2020adapterfusion} that learns to combine the information from a set of task adapters by a softmax attention layer. It learns a contextual mixture weight over adapters at layer $l$ using an attention with the softmax weights:
\begin{equation}
\setlength{\abovedisplayskip}{3pt}
\setlength{\belowdisplayskip}{3pt}
\setlength{\abovedisplayshortskip}{3pt}
\setlength{\belowdisplayshortskip}{3pt}
s_{l,k}\! =\!\operatorname{Softmax}\!\left(\!\Phi_{l,\mathcal{G}_1}\!,\!\Phi_{l,\mathcal{G}_2}\!,\!\cdots\!,\!\Phi_{l,\mathcal{G}_K};\Theta_{l,0}\!\right),
\end{equation}
where $s_{l,k}$ is used to mix the adapter outputs to be passed into the next layer, and the final layer $L$ is used to predict a task label $y$:
\begin{equation}
\setlength{\abovedisplayskip}{3pt}
\setlength{\belowdisplayskip}{3pt}
\setlength{\abovedisplayshortskip}{3pt}
\setlength{\belowdisplayshortskip}{3pt}
	y=f(\sum_{k=1}^{K} s_{L,k} \Phi_{L,\mathcal{G}_k}),
\end{equation}
where $f$ is the target task prediction head. Closely related to ours is the sparsely-gated Mixture-of-Experts layer~\cite{shazeer2017outrageously}. Alternatively, a more flexible mechanism such as {Gumbel-Softmax}~\cite{jang2016categorical} can be used for obtaining more discrete/continuous mixture weights. However, we found both alternatives underperform AdapterFusion (see Appendix for a comparison).

\section{Experiments}
\subsection{Bio-medical Knowledge Graphs}
\begin{table}[!h]
	\centering
	\resizebox{.48\textwidth}{!}{
		\begin{tabular}{lccc}
			\toprule
			\textbf{} & \textbf{\# Entities} &  \textbf{\# Relations} & \textbf{\# Triples}\\
			\midrule
			SFull & 302,332 & 229 & 4,129,726\\
			S20Rel & 263,808 & 20 & 1,750,677 \\
			\bottomrule
		\end{tabular}
	}
	\caption{Statistics of the used two SNOMED-CT KGs.}
	\label{tab:kgs}
		\vspace{-2.5mm}
\end{table}
We evaluate our proposed MoP on two KGs, named \textbf{SFull} and \textbf{S20Rel}, which are extracted from the large biomedical knowledge graph \textbf{UMLS}~\cite{bodenreider2004unified} under the \em SNOMED CT, US Edition \em vocabulary. The \textbf{SFull} KG contains the full relations and entities of SNOMED\footnote{https://www.snomed.org/}, while the \textbf{S20Rel} KG is a sub-set of \textbf{SFull} that only contains the top 20 most frequent relations. 
Note that since some relations in \textbf{SFull} are the reversed mappings of the same entity pairs, e.g. ``A has causative agent B" and ``B causative agent of A", therefore for \textbf{S20Rel} we exclude those reversed relations in the top 20 relations. Table \ref{tab:kgs} shows the statistics of the two KGs and the used 20 relations of the \textbf{S20Rel} are listed in the appendix.

%
%
\subsection{Evaluated Tasks and Datasets}

We evaluate our MoP on six datasets over various downstream tasks, including four question answering (i.e., PubMedQA, \citealt{jin2019pubmedqa}; BioAsq7b, \citealt{nentidis2019results}; BioAsq8b, \citealt{nentidis2020overview};  MedQA, \citealt{jin2020disease}), one document classification  (HoC, \citealt{baker2017initializing}), and one natural language inference (MedNLI, \citealt{romanov2018lessons}) datasets. While HoC is a multi-label classification, and MedQA is a multi-choice prediction, the rest can be formulated as a binary/multiclass classification tasks. See Appendix for the detailed description of these tasks and their datasets. 

\subsection{Pretraining with Base Models}

We experiment with three biomedical pretrained models, namely BioBERT~\cite{lee2020biobert}, SciBERT~\cite{beltagy2019scibert} and PubMedBERT~\cite{gu2020domain}, as our base models, which have shown strong progress in biomedical text mining tasks. We first partition our KGs into different number of sub-graphs (i.e. $\{5,10,20,40\}$), then for each sub-graph, we train the base models loaded with the newly initialized \textsc{Adapter} modules (with a compression rate $\text{CRate}=8$) for 1-2 epochs by minimizing the cross-entropy loss. AdamW~\cite{loshchilov2018decoupled} is used as our training optimizer, and the learning rates for all the sub-graphs are fixed to $1e-4$, as suggested by~\cite{pfeiffer2020adapterhub}. Unless specified otherwise, all the reported performances are based on a partition of 20 sub-graphs, since this was optimal for task performance (see Section \ref{subsec:num_partition} for the performances over different number of partitions.).
\begin{figure}[t]
    \centering
	\includegraphics[width = 58mm]{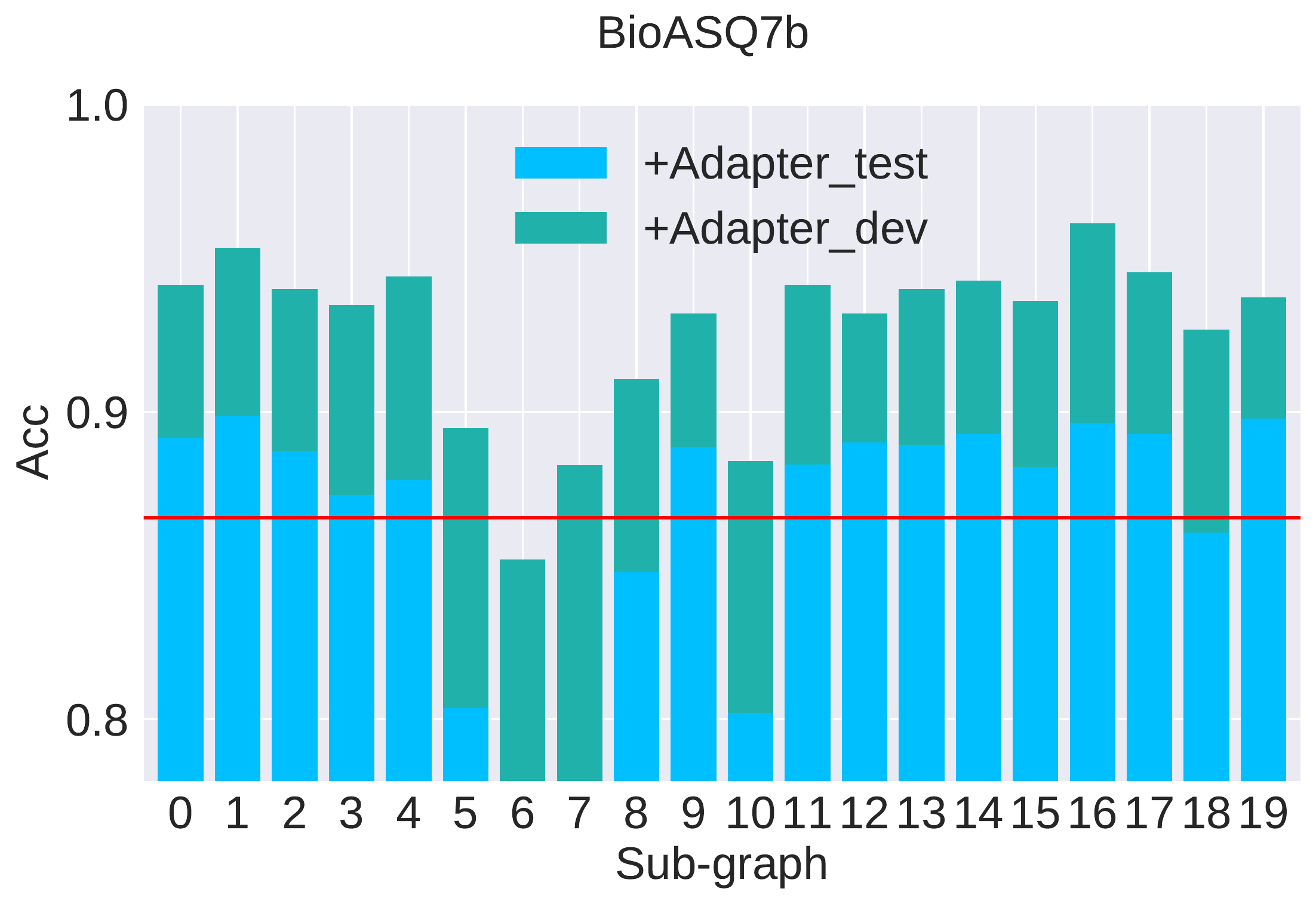}
    \hfill
    \includegraphics[width = 58mm]{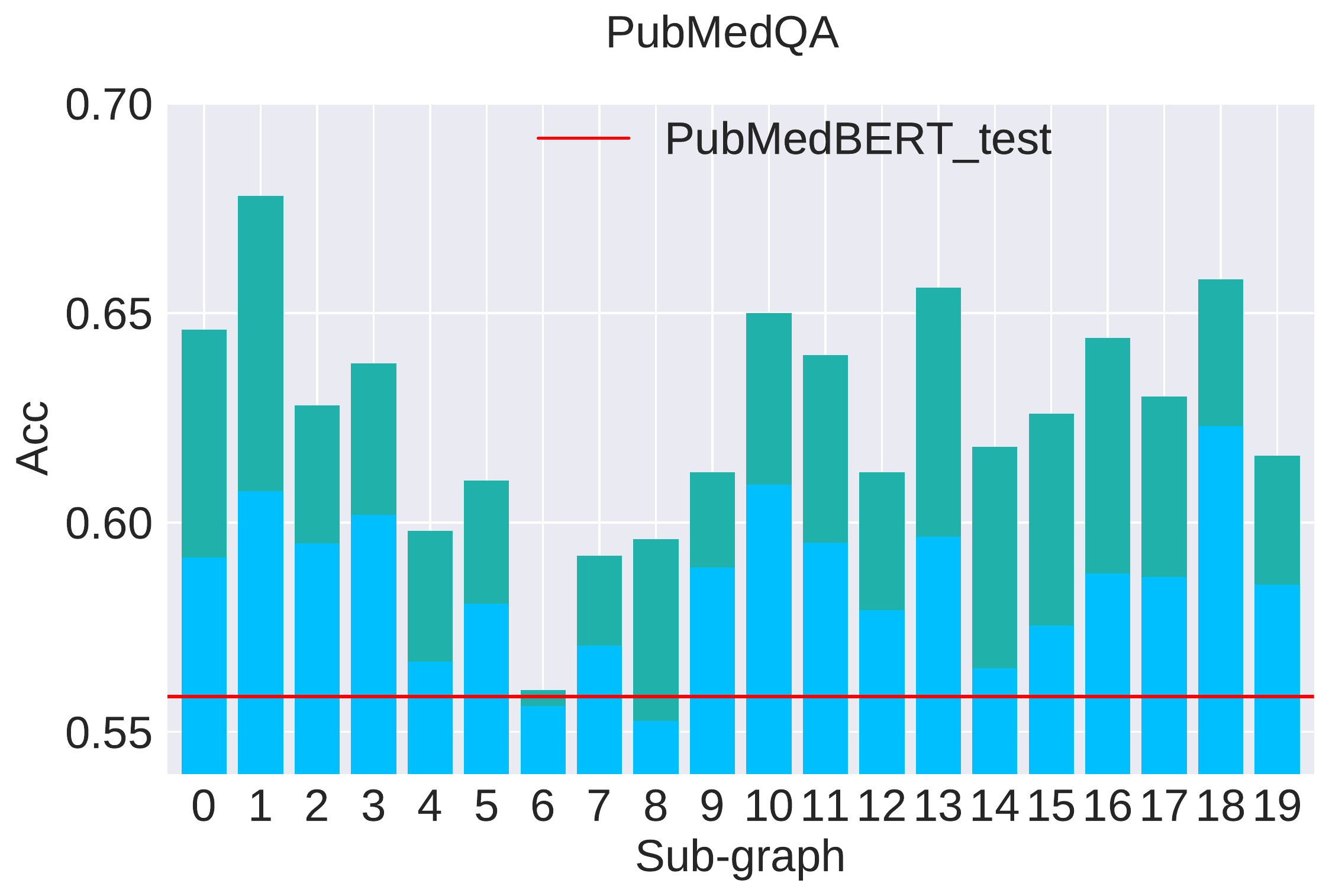}
    \caption{Performances of adapters over 20 partitions.}
    \label{fig:group_performance}
    \vspace{-2.5mm}
\end{figure}
\subsection{Partition Evaluation on Tasks} 

In Figure~\ref{fig:group_performance} we report the average performance (10 runs) of the knowledge-infused PubMedBERT on two QA datasets over partitioned \textbf{SFull}. We can see that partitions contribute to various degrees
while some (e.g. \#5) have a negligible benefit.  However, the role of the least contributing partitions could not be discarded as we repeated our downstream tasks by keeping only the top-10 performing partitions and observed the results were still worse than the model trained on all 20 partitions (i.e., accuracy drop of 2.7 on PubMedQA, and 1.4 on BioASQ7b).\footnote{See Appendix for performances on the split sub-graphs.} This highlights the importance of automatically learning the contribution weights of partitions.

\subsection{MoP Evaluation on Tasks}

Table~\ref{tab:performance} shows the overall performance of our MoP deployed on SciBERT, BioBERT and PubMedBERT pretrained models. We see that MoP pretrained on the SFull KG improves both the BioBERT and PubMedBERT models for all the tasks, while the SciBERT model can be also improved on 4 out of 6 tasks.
The result shows that MoP pretrained with the S20Rel KG achieves new SOTA performances on four tasks. This suggests  further pruning of the knowledge triples helps task performance by reducing noise, and is a promising direction to explore in future.

\begin{table}
	\centering
	\resizebox{.48\textwidth}{!}{
		\begin{tabular}{lcc}
			\toprule
			\textbf{Ratio} & \textbf{SFull} & \textbf{S20Rel} \\
			\midrule
			0\% & 2,830,674 (100.0\%) & 1,225,708 (100.0\%) \\
			10\% & 1,619,278 (57.2\%) & 990,417 (80.8\%) \\
			20\% & 1,302,096 (46.0\%) & 789,220 (64.4\%) \\
			40\% & 805,802 (28.5\%) & 479,807 (39.1\%) \\
			80\% & 289,886 (10.2\%) & 156,374 (12.8\%) \\
			100\% & 250,914 (8.9\%) & 114,695 (9.4\%) \\
			\bottomrule
		\end{tabular}
	}
	\caption{Number of training triples numbers over different shuffling ratios. With 0\% shuffling rate indicating the METIS 20-partitioned sub-graphs, and 100\% being a totally randomly generated partitions.}
	\label{tab:shuffle}
		\vspace{-2.5mm}
\end{table}

\subsection{METIS Partitioning Quality}

\begin{figure}[t]
	\centering
	\includegraphics[width = 58mm]{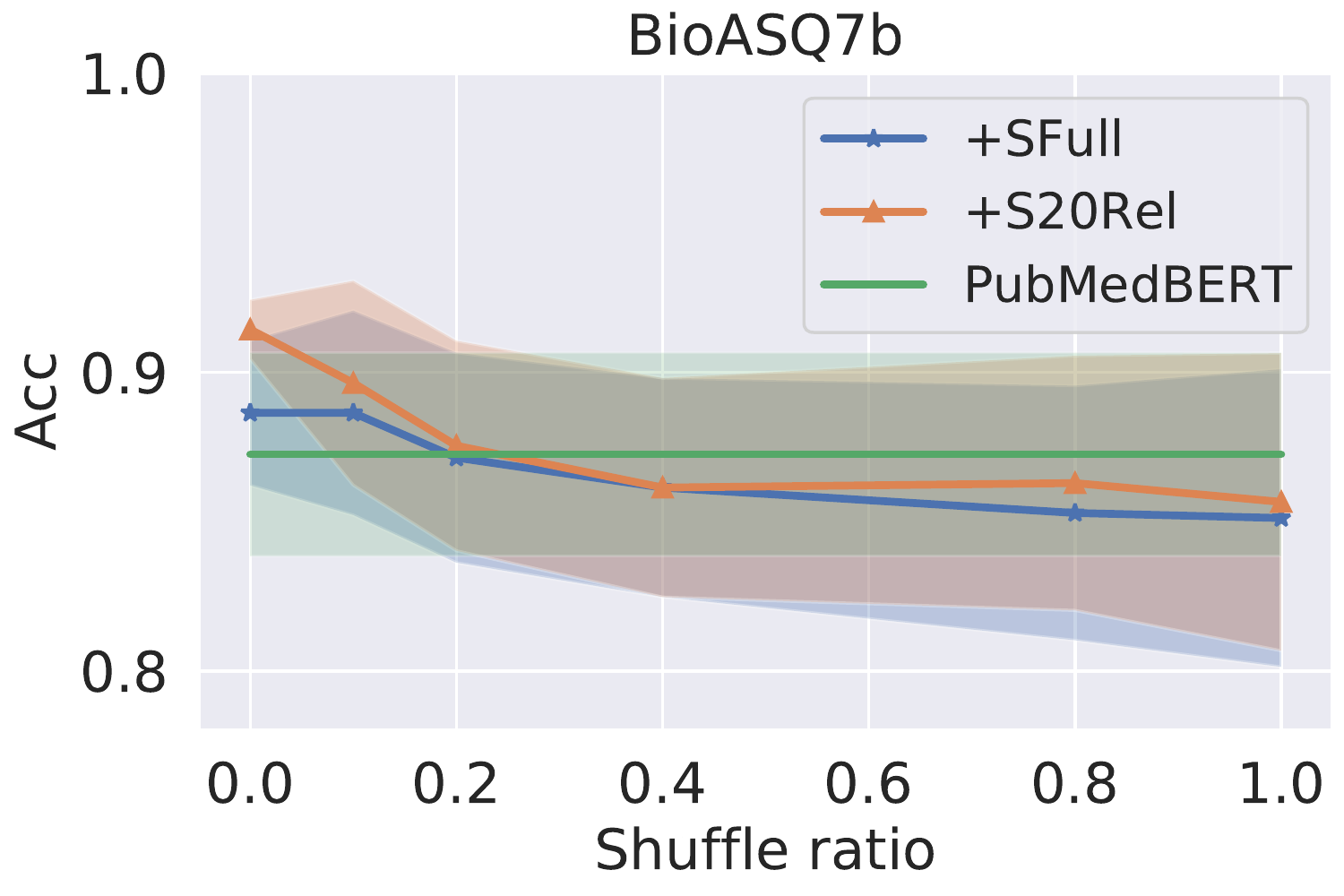}
    \includegraphics[width = 58mm]{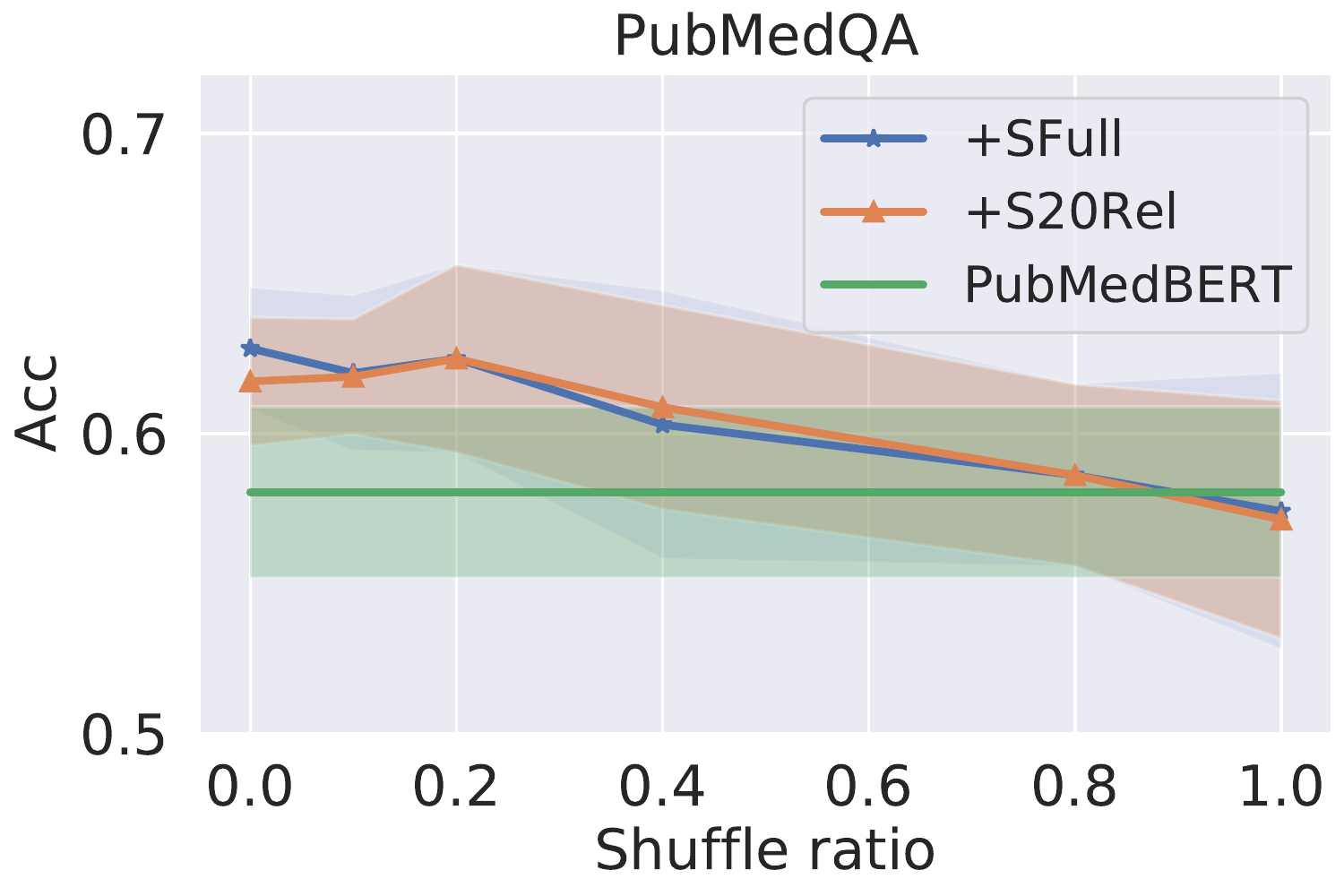}
	\caption{Performance vs. shuffling rates. The shaded regions are the standard deviations (20 runs).}
	\label{fig:shuffle}
	\vspace{-1mm}
\end{figure}

We design a controlled random partitioning scheme to test whether METIS can produce high quality partitions for training. We fix the entity size for a 20-partitioned result produced by METIS, and randomly shuffle a percentage (ranging from 0\%-100\%) of entities across all the sub-graphs. Table \ref{tab:shuffle} shows the number of training triples numbers over different shuffling ratios. In Figure~\ref{fig:shuffle} we report the results on BioASQ7b and PubMedQA under different shuffling rates. We can see that the performances of MoP on both datasets degrades significantly as the shuffling rate increases, which highlights the quality of the produced partitions.

\subsection{Performance vs. Number of Partitions} 
\label{subsec:num_partition}

Table \ref{tab:n_group} shows the performance of PubMedBERT+MoP trained on the SFull knowledge graph over different number of partitions. We can clearly see that under 20 partitions, PubMedBERT+MoP performs the best in both of the BioASQ7b and PubMedQA datasets, and an average entity size of 15k-30k for the sub-graphs usually yields better performance than others.
\begin{table}
	\centering
	\resizebox{.4\textwidth}{!}{
		\begin{tabular}{m{0.30\columnwidth}cc}
			\toprule
			\textbf{\# Sub-graphs  (Avg. \# entity)} & \textbf{BioASQ7b} & \textbf{PubMedQA} \\
			\midrule
			5 (60,466) & 86.61 & 59.22 \\
			10 (30,233) & \underline{87.86} & \underline{60.38} \\
			20 (15,116) & \textbf{88.64} & \textbf{61.74} \\
			40 (7,558) & 86.54 & 58.70 \\
			60 (5,038) & 87.11 & 59.48 \\
			\bottomrule
		\end{tabular}
	}
	\caption{Accuracy performance of MoP over different number of partitions.}
	\label{tab:n_group}
\end{table}

\subsection{Case Study} 

\begin{figure}[t]
    \centering
    \includegraphics[width = 70mm]{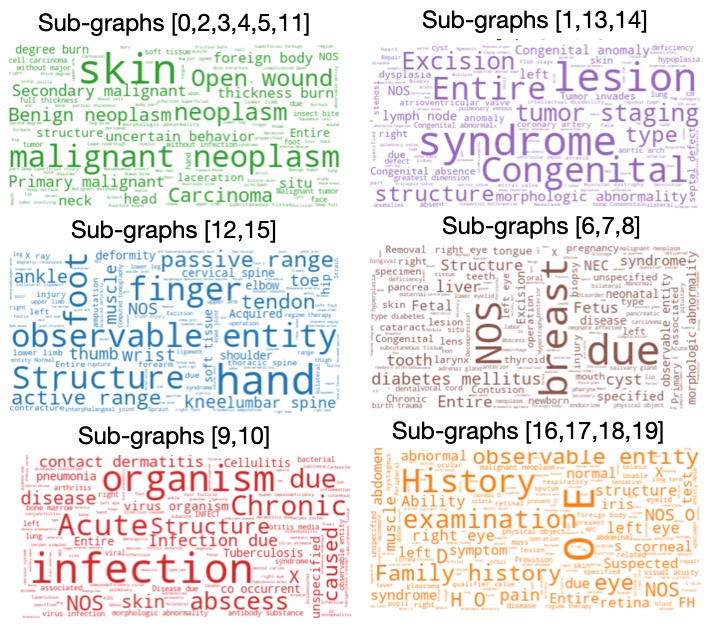}
    \includegraphics[width = 70mm]{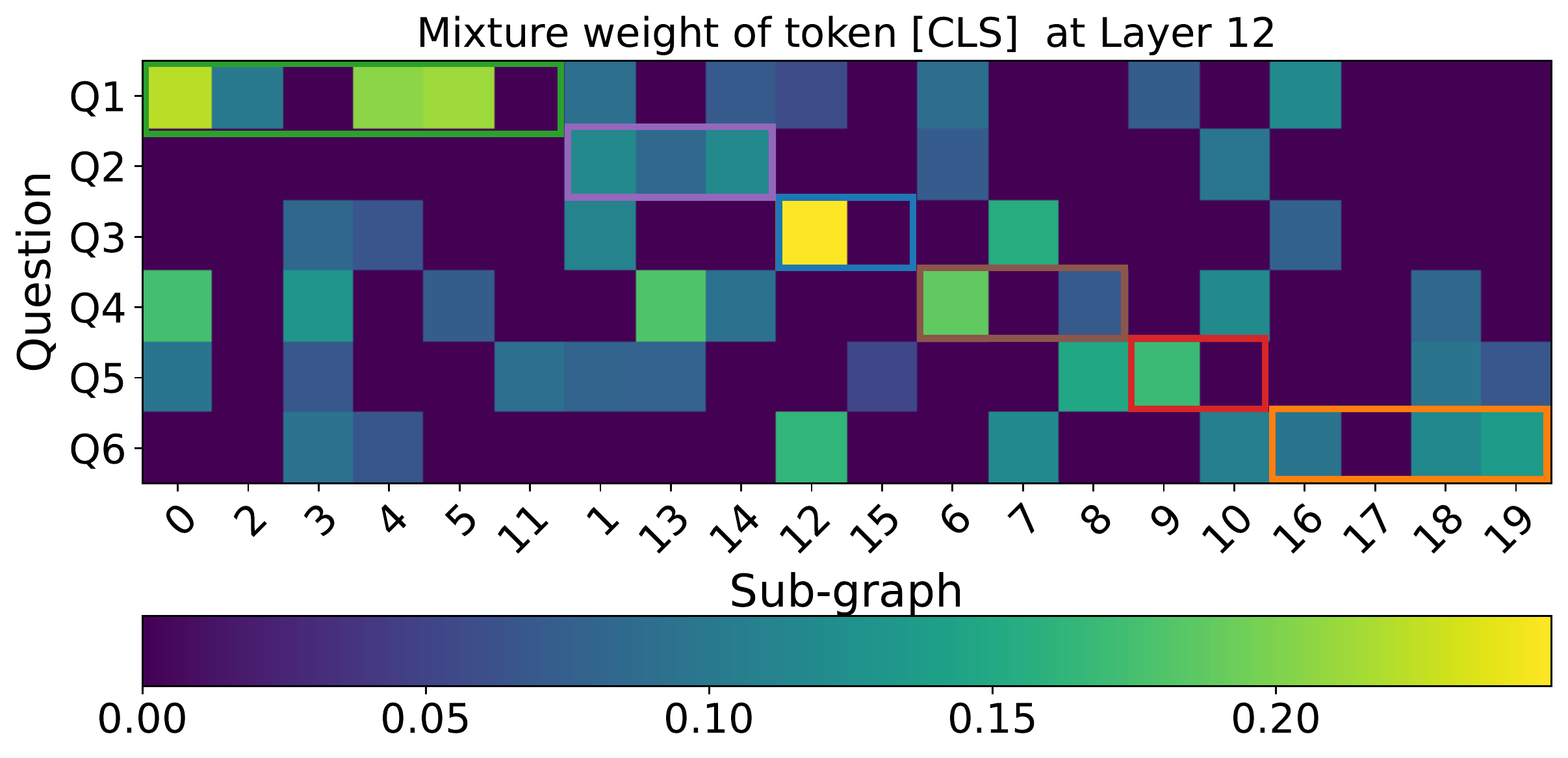}
    \resizebox{.45\textwidth}{!}{
    \begin{tabular}{ll}
	\toprule
	\color{q1} Q1 & \color{q1} Is modified vaccinia Ankara effective for smallpox? \\
	\color{q2} \underline{Q2} & \color{q2} \underline{Do de novo truncating mutations in WASF1 cause cancer?} \\
	\color{q3} Q3 & \color{q3} Thymoquinone is ineffective against radiation induced enteritis, yes or no? \\
	\color{q4} Q4 & \color{q4} Does teplizumab hold promise for diabetes prevention? \\
	\color{q5} Q5 & \color{q5} Are stem cell transplants used to treat acute kidney injury? \\
	\color{q6} Q6 & \color{q6} Are the members of the KRAB-ZNF  gene family promoting gene repression? \\
	\bottomrule
    \end{tabular}
    }
     \caption{\textbf{Top}: word cloud over 6 groups clustered from 20 sub-graphs. \textbf{Middle}: mixtures weights of MoP on 6 questions. \textbf{Bottom}: 6 questions from BioASQ8b.}
    \label{fig:word_cloud}
\end{figure}

In Figure \ref{fig:word_cloud}, we show six examples (contexts are omitted for brevity) from BioASQ8b and compare their mixture weights of the final layer [CLS] token inferred by the PubMedBERT+MoP (SFull) model. We see that each question elicits different mixture weights, indicating that MoP can leverage the expertise of different sub-graphs depending on the target example. We also plot the word cloud over six groups of sub-graphs that are clustered by k-means according to the entity name's TF-IDF feature of these sub-graphs. We can observe that MoP identifies the most related sub-graphs for each example (e.g. {\color{q2}Q2} has more weights on \textsf{\color{q2}sub-graphs [1,13,14]}, which specialise in `\textsf{\color{q2}tumor}' knowledge). This validates the effectiveness of our MoP in balancing useful knowledge across adapters.
\section{Conclusion and Future Work}
In this paper, we proposed \textbf{MoP}, a novel approach for infusing knowledge by partitioning knowledge graphs into smaller sub-graphs. We show that while the knowledge-encapsulated adapters perform very differently over different sub-graphs,  our proposed MoP can automatically leverage and balance the useful knowledge across those adapters to enhance various downstream tasks. In the future, we will evaluate our approach using some general domain KGs based on some general domain tasks. 

\section*{Acknowledgements}
Nigel Collier and Zaiqiao Meng kindly acknowledge grant-in-aid funding from ESRC (grant number ES/T012277/1).
\clearpage
\nocite{gu2020domain}
\bibliography{anthology,custom}

\begin{thebibliography}{38}
\expandafter\ifx\csname natexlab\endcsname\relax\def\natexlab#1{#1}\fi

\bibitem[{Baker and Korhonen(2017)}]{baker2017initializing}
Simon Baker and Anna Korhonen. 2017.
\newblock \href {https://www.repository.cam.ac.uk/handle/1810/285913}
  {Initializing neural networks for hierarchical multi-label text
  classification}.
\newblock \emph{BioNLP 2017}, pages 307--315.

\bibitem[{Beltagy et~al.(2019)Beltagy, Lo, and Cohan}]{beltagy2019scibert}
Iz~Beltagy, Kyle Lo, and Arman Cohan. 2019.
\newblock \href {https://www.aclweb.org/anthology/D19-1371.pdf} {{SciBERT}: A
  pretrained language model for scientific text}.
\newblock In \emph{EMNLP-IJCNLP}, pages 3606--3611.

\bibitem[{Bodenreider(2004)}]{bodenreider2004unified}
Olivier Bodenreider. 2004.
\newblock \href
  {https://academic.oup.com/nar/article/32/suppl_1/D267/2505235?login=true}
  {The unified medical language system (umls): integrating biomedical
  terminology}.
\newblock \emph{Nucleic acids research}, 32(suppl\_1):D267--D270.

\bibitem[{Chiang et~al.(2019)Chiang, Liu, Si, Li, Bengio, and
  Hsieh}]{chiang2019cluster}
Wei-Lin Chiang, Xuanqing Liu, Si~Si, Yang Li, Samy Bengio, and Cho-Jui Hsieh.
  2019.
\newblock \href {https://dl.acm.org/doi/pdf/10.1145/3292500.3330925}
  {{Cluster-GCN}: An efficient algorithm for training deep and large graph
  convolutional networks}.
\newblock In \emph{SIGKDD}, pages 257--266.

\bibitem[{De~Cao et~al.(2021)De~Cao, Izacard, Riedel, and
  Petroni}]{de2020autoregressive}
Nicola De~Cao, Gautier Izacard, Sebastian Riedel, and Fabio Petroni. 2021.
\newblock \href {https://arxiv.org/pdf/2010.00904.pdf} {Autoregressive entity
  retrieval}.
\newblock In \emph{ICLR}.

\bibitem[{Defferrard et~al.(2016)Defferrard, Bresson, and
  Vandergheynst}]{defferrard2016convolutional}
Micha{\"e}l Defferrard, Xavier Bresson, and Pierre Vandergheynst. 2016.
\newblock \href {https://arxiv.org/pdf/1606.09375.pdf} {Convolutional neural
  networks on graphs with fast localized spectral filtering}.
\newblock In \emph{NeurIPS}, pages 3844--3852.

\bibitem[{Goodwin and Demner-Fushman(2020)}]{goodwin2020enhancing}
Travis~R Goodwin and Dina Demner-Fushman. 2020.
\newblock \href {https://www.ncbi.nlm.nih.gov/pmc/articles/PMC7757122/}
  {Enhancing question answering by injecting ontological knowledge through
  regularization}.
\newblock In \emph{EMNLP}, volume 2020, page~56.

\bibitem[{Gu et~al.(2020)Gu, Tinn, Cheng, Lucas, Usuyama, Liu, Naumann, Gao,
  and Poon}]{gu2020domain}
Yu~Gu, Robert Tinn, Hao Cheng, Michael Lucas, Naoto Usuyama, Xiaodong Liu,
  Tristan Naumann, Jianfeng Gao, and Hoifung Poon. 2020.
\newblock \href {https://arxiv.org/pdf/2007.15779} {Domain-specific language
  model pretraining for biomedical natural language processing}.
\newblock \emph{arXiv preprint arXiv:2007.15779}.

\bibitem[{He et~al.(2020)He, Zhu, Zhang, Chen, and Caverlee}]{he2020infusing}
Yun He, Ziwei Zhu, Yin Zhang, Qin Chen, and James Caverlee. 2020.
\newblock \href {https://arxiv.org/pdf/2010.03746.pdf} {Infusing disease
  knowledge into {BERT} for health question answering, medical inference and
  disease name recognition}.
\newblock In \emph{EMNLP}, pages 4604--4614.

\bibitem[{Hinton et~al.(2015)Hinton, Vinyals, and Dean}]{hinton2015distilling}
Geoffrey Hinton, Oriol Vinyals, and Jeff Dean. 2015.
\newblock \href {https://arxiv.org/pdf/1503.02531.pdf} {Distilling the
  knowledge in a neural network}.
\newblock In \emph{NeurIPS}.

\bibitem[{Houlsby et~al.(2019)Houlsby, Giurgiu, Jastrzebski, Morrone,
  De~Laroussilhe, Gesmundo, Attariyan, and Gelly}]{houlsby2019parameter}
Neil Houlsby, Andrei Giurgiu, Stanislaw Jastrzebski, Bruna Morrone, Quentin
  De~Laroussilhe, Andrea Gesmundo, Mona Attariyan, and Sylvain Gelly. 2019.
\newblock \href {http://proceedings.mlr.press/v97/houlsby19a.html}
  {Parameter-efficient transfer learning for nlp}.
\newblock In \emph{ICML}, pages 2790--2799.

\bibitem[{Jang et~al.(2017)Jang, Gu, and Poole}]{jang2016categorical}
Eric Jang, Shixiang Gu, and Ben Poole. 2017.
\newblock \href {https://arxiv.org/pdf/1611.01144.pdf} {Categorical
  reparameterization with gumbel-softmax}.
\newblock In \emph{ICLR}.

\bibitem[{Jin et~al.(2020)Jin, Pan, Oufattole, Weng, Fang, and
  Szolovits}]{jin2020disease}
Di~Jin, Eileen Pan, Nassim Oufattole, Wei-Hung Weng, Hanyi Fang, and Peter
  Szolovits. 2020.
\newblock \href {https://arxiv.org/pdf/2009.13081.pdf} {What disease does this
  patient have? a large-scale open domain question answering dataset from
  medical exams}.
\newblock \emph{arXiv preprint arXiv:2009.13081}.

\bibitem[{Jin et~al.(2019)Jin, Dhingra, Liu, Cohen, and Lu}]{jin2019pubmedqa}
Qiao Jin, Bhuwan Dhingra, Zhengping Liu, William Cohen, and Xinghua Lu. 2019.
\newblock \href {https://arxiv.org/pdf/1909.06146.pdf} {{PubMedQA}: A dataset
  for biomedical research question answering}.
\newblock In \emph{EMNLP-IJCNLP}, pages 2567--2577.

\bibitem[{Karypis and Kumar(1998)}]{karypis1998fast}
George Karypis and Vipin Kumar. 1998.
\newblock \href {https://epubs.siam.org/doi/pdf/10.1137/S1064827595287997} {A
  fast and high quality multilevel scheme for partitioning irregular graphs}.
\newblock \emph{SIAM Journal on scientific Computing}, 20(1):359--392.

\bibitem[{Lee et~al.(2020)Lee, Yoon, Kim, Kim, Kim, So, and
  Kang}]{lee2020biobert}
Jinhyuk Lee, Wonjin Yoon, Sungdong Kim, Donghyeon Kim, Sunkyu Kim, Chan~Ho So,
  and Jaewoo Kang. 2020.
\newblock \href
  {https://academic.oup.com/bioinformatics/article/36/4/1234/5566506?login=true}
  {{BioBERT}: a pre-trained biomedical language representation model for
  biomedical text mining}.
\newblock \emph{Bioinformatics}, 36(4):1234--1240.

\bibitem[{Li et~al.(2020)Li, Hu, Chen, Peng, and Wang}]{li2020towards}
Dongfang Li, Baotian Hu, Qingcai Chen, Weihua Peng, and Anqi Wang. 2020.
\newblock \href {https://www.aclweb.org/anthology/2020.emnlp-main.111.pdf}
  {Towards medical machine reading comprehension with structural knowledge and
  plain text}.
\newblock In \emph{EMNLP}, pages 1427--1438.

\bibitem[{Liu et~al.(2021)Liu, Shareghi, Meng, Basaldella, and
  Collier}]{liu2020self}
Fangyu Liu, Ehsan Shareghi, Zaiqiao Meng, Marco Basaldella, and Nigel Collier.
  2021.
\newblock \href {https://arxiv.org/pdf/2010.11784.pdf} {Self-alignment
  pre-training for biomedical entity representations}.
\newblock \emph{NAACL}.

\bibitem[{Loshchilov and Hutter(2018)}]{loshchilov2018decoupled}
Ilya Loshchilov and Frank Hutter. 2018.
\newblock \href {https://arxiv.org/pdf/1711.05101.pdf} {Decoupled weight decay
  regularization}.
\newblock In \emph{ICLR}.

\bibitem[{Mikolov et~al.(2013)Mikolov, Sutskever, Chen, Corrado, and
  Dean}]{mikolov2013distributed}
Tomas Mikolov, Ilya Sutskever, Kai Chen, Greg Corrado, and Jeffrey Dean. 2013.
\newblock \href {https://arxiv.org/pdf/1310.4546.pdf} {Distributed
  representations of words and phrases and their compositionality}.
\newblock In \emph{NeurIPS}, pages 3111--3119.

\bibitem[{Nentidis et~al.(2019)Nentidis, Bougiatiotis, Krithara, and
  Paliouras}]{nentidis2019results}
Anastasios Nentidis, Konstantinos Bougiatiotis, Anastasia Krithara, and
  Georgios Paliouras. 2019.
\newblock \href
  {https://link.springer.com/chapter/10.1007/978-3-030-43887-6_51} {Results of
  the seventh edition of the {BioASQ} challenge}.
\newblock In \emph{ECML PKDD}, pages 553--568.

\bibitem[{Nentidis et~al.(2020)Nentidis, Krithara, Bougiatiotis, Krallinger,
  Rodriguez-Penagos, Villegas, and Paliouras}]{nentidis2020overview}
Anastasios Nentidis, Anastasia Krithara, Konstantinos Bougiatiotis, Martin
  Krallinger, Carlos Rodriguez-Penagos, Marta Villegas, and Georgios Paliouras.
  2020.
\newblock \href
  {https://link.springer.com/chapter/10.1007/978-3-030-58219-7_16} {Overview of
  {BioASQ} 2020: The eighth {BioASQ} challenge on large-scale biomedical
  semantic indexing and question answering}.
\newblock In \emph{CLEF}, pages 194--214.

\bibitem[{Peng et~al.(2019)Peng, Yan, and Lu}]{peng2019transfer}
Yifan Peng, Shankai Yan, and Zhiyong Lu. 2019.
\newblock \href {https://arxiv.org/pdf/1906.05474.pdf} {Transfer learning in
  biomedical natural language processing: An evaluation of {BERT} and {ELMo} on
  ten benchmarking datasets}.
\newblock In \emph{Proceedings of the 18th BioNLP Workshop and Shared Task},
  pages 58--65.

\bibitem[{Peters et~al.(2019)Peters, Neumann, Logan, Schwartz, Joshi, Singh,
  and Smith}]{peters2019knowledge}
Matthew~E Peters, Mark Neumann, Robert Logan, Roy Schwartz, Vidur Joshi, Sameer
  Singh, and Noah~A Smith. 2019.
\newblock \href {https://arxiv.org/pdf/1909.04164.pdf} {Knowledge enhanced
  contextual word representations}.
\newblock In \emph{EMNLP-IJCNLP}, pages 43--54.

\bibitem[{Petroni et~al.(2021)Petroni, Piktus, Fan, Lewis, Yazdani, De~Cao,
  Thorne, Jernite, Karpukhin, Maillard et~al.}]{petroni2020kilt}
Fabio Petroni, Aleksandra Piktus, Angela Fan, Patrick Lewis, Majid Yazdani,
  Nicola De~Cao, James Thorne, Yacine Jernite, Vladimir Karpukhin, Jean
  Maillard, et~al. 2021.
\newblock \href {https://arxiv.org/pdf/2009.02252.pdf} {{KILT}: a benchmark for
  knowledge intensive language tasks}.
\newblock \emph{NAACL}.

\bibitem[{Pfeiffer et~al.(2020{\natexlab{a}})Pfeiffer, Kamath, R{\"u}ckl{\'e},
  Cho, and Gurevych}]{pfeiffer2020adapterfusion}
Jonas Pfeiffer, Aishwarya Kamath, Andreas R{\"u}ckl{\'e}, Kyunghyun Cho, and
  Iryna Gurevych. 2020{\natexlab{a}}.
\newblock \href {https://arxiv.org/pdf/2005.00247.pdf} {Adapterfusion:
  Non-destructive task composition for transfer learning}.
\newblock In \emph{EMNLP}.

\bibitem[{Pfeiffer et~al.(2020{\natexlab{b}})Pfeiffer, R{\"u}ckl{\'e}, Poth,
  Kamath, Vuli{\'c}, Ruder, Cho, and Gurevych}]{pfeiffer2020adapterhub}
Jonas Pfeiffer, Andreas R{\"u}ckl{\'e}, Clifton Poth, Aishwarya Kamath, Ivan
  Vuli{\'c}, Sebastian Ruder, Kyunghyun Cho, and Iryna Gurevych.
  2020{\natexlab{b}}.
\newblock \href {https://arxiv.org/pdf/2007.07779.pdf} {{AdapterHub}: A
  framework for adapting transformers}.
\newblock In \emph{EMNLP}, pages 46--54.

\bibitem[{Romanov and Shivade(2018)}]{romanov2018lessons}
Alexey Romanov and Chaitanya Shivade. 2018.
\newblock \href {https://arxiv.org/pdf/1808.06752.pdf} {Lessons from natural
  language inference in the clinical domain}.
\newblock In \emph{EMNLP}, pages 1586--1596.

\bibitem[{Shazeer et~al.(2017)Shazeer, Mirhoseini, Maziarz, Davis, Le, Hinton,
  and Dean}]{shazeer2017outrageously}
Noam Shazeer, Azalia Mirhoseini, Krzysztof Maziarz, Andy Davis, Quoc Le,
  Geoffrey Hinton, and Jeff Dean. 2017.
\newblock \href {https://arxiv.org/pdf/1701.06538.pdf} {Outrageously large
  neural networks: The sparsely-gated mixture-of-experts layer}.
\newblock In \emph{ICLR}.

\bibitem[{Sun et~al.(2020)Sun, Shao, Qiu, Guo, Hu, Huang, and
  Zhang}]{sun2020colake}
Tianxiang Sun, Yunfan Shao, Xipeng Qiu, Qipeng Guo, Yaru Hu, Xuan-Jing Huang,
  and Zheng Zhang. 2020.
\newblock \href {https://arxiv.org/pdf/2010.00309.pdf} {Colake: Contextualized
  language and knowledge embedding}.
\newblock In \emph{COLING}, pages 3660--3670.

\bibitem[{Sun et~al.(2019)Sun, Wang, Li, Feng, Chen, Zhang, Tian, Zhu, Tian,
  and Wu}]{sun2019ernie}
Yu~Sun, Shuohuan Wang, Yukun Li, Shikun Feng, Xuyi Chen, Han Zhang, Xin Tian,
  Danxiang Zhu, Hao Tian, and Hua Wu. 2019.
\newblock \href {https://arxiv.org/pdf/1904.09223} {Ernie: Enhanced
  representation through knowledge integration}.
\newblock \emph{arXiv preprint arXiv:1904.09223}.

\bibitem[{Wang et~al.(2020)Wang, Tang, Duan, Wei, Huang, Cao, Jiang, Zhou
  et~al.}]{wang2020k}
Ruize Wang, Duyu Tang, Nan Duan, Zhongyu Wei, Xuanjing Huang, Cuihong Cao,
  Daxin Jiang, Ming Zhou, et~al. 2020.
\newblock \href {https://arxiv.org/pdf/2002.01808.pdf} {K-adapter: Infusing
  knowledge into pre-trained models with adapters}.
\newblock \emph{arXiv preprint arXiv:2002.01808}.

\bibitem[{Wang et~al.(2021)Wang, Gao, Zhu, Zhang, Liu, Li, and
  Tang}]{wang2019kepler}
Xiaozhi Wang, Tianyu Gao, Zhaocheng Zhu, Zhengyan Zhang, Zhiyuan Liu, Juanzi
  Li, and Jian Tang. 2021.
\newblock \href
  {https://direct.mit.edu/tacl/article/doi/10.1162/tacl_a_00360/98089}
  {{KEPLER}: A unified model for knowledge embedding and pre-trained language
  representation}.
\newblock \emph{Transactions of the Association for Computational Linguistics},
  9:176--194.

\bibitem[{Xiong et~al.(2019)Xiong, Du, Wang, and
  Stoyanov}]{xiong2019pretrained}
Wenhan Xiong, Jingfei Du, William~Yang Wang, and Veselin Stoyanov. 2019.
\newblock \href {https://arxiv.org/pdf/1912.09637.pdf} {Pretrained
  encyclopedia: Weakly supervised knowledge-pretrained language model}.
\newblock In \emph{ICLR}.

\bibitem[{Yu et~al.(2020)Yu, Zhu, Yang, and Zeng}]{yu2020jaket}
Donghan Yu, Chenguang Zhu, Yiming Yang, and Michael Zeng. 2020.
\newblock \href {https://arxiv.org/pdf/2010.00796.pdf} {Jaket: Joint
  pre-training of knowledge graph and language understanding}.
\newblock \emph{arXiv preprint arXiv:2010.00796}.

\bibitem[{Yuan et~al.(2021)Yuan, Liu, Tan, Huang, and
  Huang}]{yuan2021improving}
Zheng Yuan, Yijia Liu, Chuanqi Tan, Songfang Huang, and Fei Huang. 2021.
\newblock \href {https://arxiv.org/pdf/2104.10344.pdf} {Improving biomedical
  pretrained language models with knowledge}.
\newblock \emph{arXiv preprint arXiv:2104.10344}.

\bibitem[{Zhang et~al.(2019)Zhang, Han, Liu, Jiang, Sun, and
  Liu}]{zhang2019ernie}
Zhengyan Zhang, Xu~Han, Zhiyuan Liu, Xin Jiang, Maosong Sun, and Qun Liu. 2019.
\newblock \href {https://arxiv.org/pdf/1905.07129.pdf} {Ernie: Enhanced
  language representation with informative entities}.
\newblock In \emph{ACL}, pages 1441--1451.

\bibitem[{Zheng et~al.(2020)Zheng, Ma, Wang, Zhou, Su, Song, Gan, Zhang, and
  Karypis}]{zheng2020distdgl}
Da~Zheng, Chao Ma, Minjie Wang, Jinjing Zhou, Qidong Su, Xiang Song, Quan Gan,
  Zheng Zhang, and George Karypis. 2020.
\newblock \href {https://arxiv.org/pdf/2010.05337.pdf} {{DistDGL}: Distributed
  graph neural network training for billion-scale graphs}.
\newblock \emph{arXiv preprint arXiv:2010.05337}.

\end{thebibliography}
\bibliographystyle{acl_natbib}
\clearpage
\renewcommand{\thesection}{\Alph{section}}
\renewcommand{\thesubsection}{A.\arabic{subsection}}
\section*{Appendix}
\subsection{The used 20 relations in the \textbf{S20Rel} knowledge graph}

\begin{table}[!h]
	\centering
	\resizebox{.35\textwidth}{!}{
		\begin{tabular}{cc}
			\toprule
			\textbf{ID} & \textbf{Relation} \\
			\midrule
			0 & has causative agent \\
            1 & has active ingredient \\
            2 & has direct substance \\
            3 & has finding site \\
            4 & has direct procedure site \\
            5 & has procedure site \\
            6 & possibly equivalent to \\
            7 & has occurrence \\
            8 & has associated morphology \\
            9 & has direct morphology \\
            10 & interprets \\
            11 & has method \\
            12 & has direct device \\
            13 & has dose form \\
            14 & has subject relationship context \\
            15 & has pathological process \\
            16 & has interpretation \\
            17 & moved to \\
            18 & has intent \\
            19 & has temporal context \\
			\bottomrule
		\end{tabular}
	}
	\caption{Relations used in \textbf{S20Rel} knowledge graph.}
	\label{tab:20rel}
\end{table}

\subsection{Evaluated Datasets and Experiment details}
We evaluate our MoP on six datasets over various downstream tasks, including four question answering (i.e., PubMedQA, \citealt{jin2019pubmedqa}; BioAsq7b, \citealt{nentidis2019results}; BioAsq8b, \citealt{nentidis2020overview};  MedQA, \citealt{jin2020disease}), one document classification  (HoC, \citealt{baker2017initializing}), and one natural language inference (MedNLI, \citealt{romanov2018lessons}) datasets. While HoC is a multi-label classification, and MedQA is a multi-choice prediction, the rest can be formulated as a binary/multiclass classification tasks. 
\label{sec:dataset}
\begin{itemize}
	\item \textbf{HoC}~\cite{baker2017initializing}: The Hallmarks of Cancer corpus was extracted from 1852 PubMed publication abstracts by ~\citet{baker2017initializing}, and the class labels were manually annotated by experts according to the Hallmarks of Cancer taxonomy. The taxonomy consists of 37 classes in a hierarchy, but in this paper we only consider the ten top-level ones. We use the publicly available train/dev/test split created by ~\cite{gu2020domain} and report the average performance over five runs by the average micro F1 across the ten cancer hallmarks.
	\item \textbf{PubMedQA}~\cite{jin2019pubmedqa}: This is a question answering dataset that contains a set of research questions, each with a reference text from a PubMed abstract as well as an annotated label of whether the text contains the answer to the research question (\em yes/maybe/no \em). We use the original \em train/dev/test \em split with 450/50/500 questions, respectively. The reported performance are the average of ten runs under the accuracy metric.
	\item \textbf{BioASQ7b, BioASQ8b}~\cite{nentidis2019results,nentidis2020overview}: The both BioASQ datasets are \em yes/no \em question answering tasks annotated by biomedical experts. Each question is paired with a reference text containing multiple sentences from a PubMed abstract and a \em yes/no \em answer. We use the official \em train/dev/test \em splits, i.e. 670/75/140 and 729/152/152 for BioASQ7b and BioASQ8b respectively, and the reported performances are the average of ten runs under the accuracy metric.
	\item \textbf{MedNLI}~\cite{romanov2018lessons}: MedNLI is a Natural Language Inference  (NLI) collection of sentence pairs extracted from MIMIC-III, a large clinical database. The objective of the NLI task is to determine if a given hypothesis can be inferred from a given premise. This task  is formulated as the document classification task over three labels: \{entailment, contradiction,neutral\}. We use the same train/dev/test split generated by \citet{romanov2018lessons}, and report the average accuracy performance over three runs.
	\item \textbf{MedQA}~\cite{jin2020disease}: MedQA is a publicly available large-scale multiple-choice question answering dataset extracted from the professional medical board exams. It covers three languages: English, simplified Chinese, and traditional Chinese, but in this paper we only adopt the English set, which is split by \citet{jin2020disease}. Following~\cite{jin2020disease}, we use the Elasticsearch system to retrieve the top 25 sentences to each question+choice pair as the context for each choice, and concatenate them to obtain the normalized log probability over the five choices. Since this dataset is very large, we only report the average accuracy performance under three runs for all the models.
\end{itemize}

\subsection{Comparison of Different Mixture Approaches}
Our MoP approach first infuses the factual knowledge for all the partitioned sub-graphs using their respective adapters with newly initialized parameters, then these knowledge-encapsulated adapters are further fine-tuned alone with the underlying BERT model through mixture layers. In this paper, we explored three approaches for implementing the mixture layers, which are described as follows:
\begin{itemize}
    \item \textbf{Softmax.} As the default mixture layers deployed in our MoP, \textbf{AdapterFusion} is a recent proposed  model~\cite{pfeiffer2020adapterfusion} that learns to combine the information from a set of tasks adapters by a softmax attention layer. In particular, the outputs from different adapters at layer $l$ are combined using a contextual mixture weight calculated by a softmax  over these adapters:
    \begin{equation}
    s_{l,k}\! =\!\operatorname{Softmax}\!\left(\!\Phi_{l,\mathcal{G}_1}\!,\!\Phi_{l,\mathcal{G}_2}\!,\!\cdots\!,\!\Phi_{l,\mathcal{G}_K};\Theta_{l,0}\!\right).
    \end{equation}
    For brevity, we denote our MoP with the original AdapterFusion mixture layers as \textbf{Softmax}.
    \item \textbf{Gumbel.} We also extend the AdapterFusion by replacing the softmax layer with the \textbf{Gumbel-Softmax}~\cite{jang2016categorical} layer for obtaining more discrete mixture weights:
    {\small 
\begin{align}
    s_{l,k} \!&=\! \text{Gumbel-Softmax}\!\left(\!\Phi_{l,\mathcal{G}_1}\!,\!\Phi_{l,\mathcal{G}_2}\!,\!\cdots\!,\!\Phi_{l,\mathcal{G}_K};\Theta_{l,0}\!\right) \nonumber
    \\&= \frac{\exp \left(\log \Phi_{l,\mathcal{G}_k}+g_{k}\right) / \tau}{\sum_{i=1}^{K} \exp \left(\log \Phi_{l,\mathcal{G}_i}+g_{i}\right) / \tau},
    \end{align}
    }%
    where $g_1,\cdots,g_K$ are i.i.d samples drawn from $Gumbel(0,1)$ distribution, and $\tau$ is a hyper-parameter controlling the discreteness. For brevity, we denote our MoP with the Gumbel-Softmax AdapterFusion mixture layers as \textbf{Gumbel}.
    \item \textbf{MoE.} \textbf{Mix-of-Experts (MoE)} is a type of general purpose neural network component for selecting a combination of the experts to process each input. In particular, we use the the sparsely-gated mixture-of-experts, introduced by~\citet{shazeer2017outrageously}, for obtaining a top-K sparse mixture of these adapters. And the mixture weights are calculated by:
    \begin{align}
    &s_{l,k}=\operatorname{Softmax}(\operatorname{TopK}(H(\Phi_{l,\mathcal{G}_k}), K)),
    \end{align}
    where $H(\Phi_{l,\mathcal{G}_k})$ is a function for transferring hidden variables into scalars with tunable Gaussian noise, and $\operatorname{TopK}(\cdot)$ is a function for keeping only the top $K$ values. We denote our MoP with this mixture approach as \textbf{MoE}.
\end{itemize}
Table \ref{tab:mix_layer} shows the performance comparison of the three mixture approaches on the BioASQ7b and PubMedQA tasks. Note that Gumbel and MoE have additional hyper-parameters, i.e. $\tau$ and $K$, for controlling the discreteness and topK respectively. From Table \ref{tab:mix_layer}, we can see that the original AdapterFusion with the softmax layer outperforms other two mixture approaches on all the hyper-parameter choices. This result justifies the choice of the mixture layers in our MoP model.

\begin{table}
	\centering
	\resizebox{.38\textwidth}{!}{
		\begin{tabular}{lcc}
			\toprule
			\textbf{Model} & \textbf{BioASQ7b} & \textbf{PubMedQA} \\
			\midrule
			Gumbel ($\tau=0.1$) & 85.21$\pm7.5$ & {59.90}$\pm2.8$ \\
			Gumbel ($\tau=0.25$) & 87.07$\pm2.7$ & {59.94}$\pm1.8$ \\
			Gumbel ($\tau=0.5$) & 87.07$\pm3.8$ & {59.46}$\pm2.1$ \\
			Gumbel ($\tau=0.75$) & 87.14$\pm3.6$ & {59.35}$\pm1.9$\\
			MoE ($K=2$) & 86.93$\pm6.5$ & 60.06$\pm5.1$  \\
			MoE ($K=3$) & 87.36$\pm6.2$ & 60.26$\pm4.5$ \\
			MoE ($K=5$) & 86.79$\pm4.5$ &  59.32$\pm4.4$\\
			MoE ($K=10$) & 85.57$\pm4.8$ & 59.02$\pm5.7$\\
			MoE ($K=15$) & 88.29$\pm2.3$ & 61.52$\pm4.9$ \\
			Softmax & \textbf{88.64}$\pm3.0$ & \textbf{61.74}$\pm2.7$ \\
			\bottomrule
		\end{tabular}
	}
	\caption{Performance comparison on BioASQ7b and PubMedQA tasks over the three mixture approaches. }
	\label{tab:mix_layer}
\end{table}

\subsection{Performance on the Split Sub-graphs}

To further validate that the performance improvements of the evaluated BERTs using our MoP are gained due to the infused knowledge from the sub-graph adapters, rather than the newly added more parameters of adapters, we split the partitioned sub-graphs into two groups according to their test performance ranking, and use our MoP to fine-tune the adapters of each group. Table \ref{tab:biasq_split_group} and \ref{tab:pubmedqa_split_group} show the performances of all the adapters and our MoP combining the grouped adapters over train/dev/test sets of the BioASQ7b and PubMedQA datasets respectively. As we can see from the two tables, our MoP fine-tuned under the group of higher performance adapters can consistently obtain better performance than the group of the lower performance adapters. Note that we have shown that in Figure \ref{fig:group_performance} adapters pretrained by different sub-graphs show quite different performances, and each sub-graph adapter can capture different factual knowledge. Then, the result from Tab \ref{tab:biasq_split_group} and \ref{tab:pubmedqa_split_group} further validates that the marginal performance gains of our MoP are indeed caused by the mixture of knowledge infused from adapters.

\begin{table}[htp]
	\centering
	\includegraphics[width = 70mm]{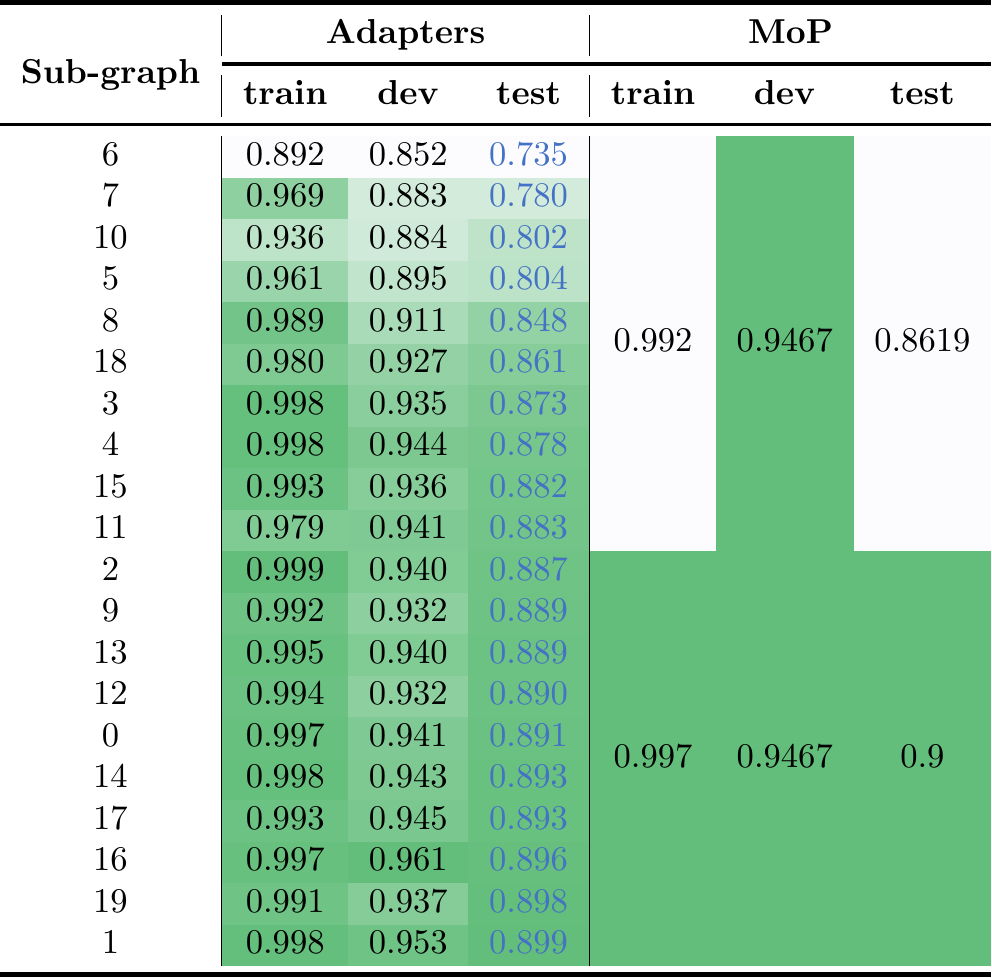}
	\caption{MoP Performance on BioASQ7b over two groups of sub-graphs, which are divided according to their test accuracy performance.}
	\label{tab:biasq_split_group}
\end{table}

\begin{table}[htp]
	\centering
	\includegraphics[width = 70mm]{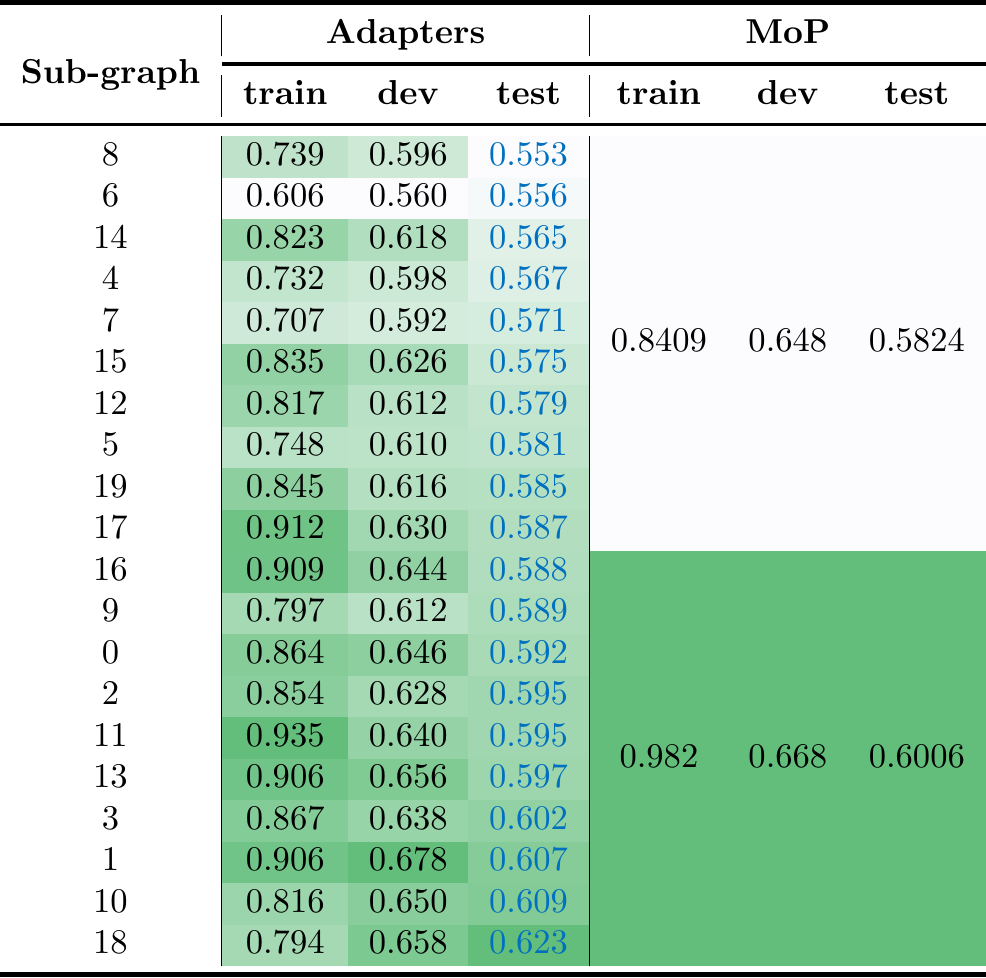}
	\caption{MoP Performance on PubMedQA over two groups of sub-graphs, which are divided according to their test accuracy performance.}
	\label{tab:pubmedqa_split_group}
\end{table}

\end{document}